\documentclass{article}

\usepackage[preprint]{neurips_2024}

\usepackage[utf8]{inputenc} 
\usepackage[T1]{fontenc}    
\usepackage{hyperref}       
\usepackage{url}            
\usepackage{booktabs}       
\usepackage{amsfonts}       
\usepackage{nicefrac}       
\usepackage{microtype}      
\usepackage{xcolor}         
\usepackage{amsmath}
\usepackage{graphicx}
\usepackage{booktabs}       
\usepackage{siunitx}        
\usepackage{multirow}       
\usepackage{subcaption}

\title{Conformal Sets in Multiple-Choice Question Answering under Black-Box Settings with Provable Coverage Guarantees}

\author{%
  Guang Yang\thanks{Use footnote for providing further information
    about author (webpage, alternative address)---\emph{not} for acknowledging
    funding agencies.} \\
  School of Electrical Engineering\\
  University of Jinan\\
  Jinan, Shandong 250024 \\
  \texttt{guangyang.acad@gmail.com} \\
  \And
  XinYangLiu \\
  School of Electrical Engineering \\
  University of Jinan\\
  Jinan, Shandong 250024 \\
  \texttt{liuliuxinyang09@gmail.com} \\
}

\begin{document}

\maketitle

\begin{abstract}
Large Language Models (LLMs) have shown remarkable progress in multiple-choice question answering (MCQA), but their inherent unreliability, such as hallucination and overconfidence, limits their application in high-risk domains. To address this, we propose a frequency-based uncertainty quantification method under black-box settings, leveraging conformal prediction (CP) to ensure provable coverage guarantees. Our approach involves multiple independent samplings of the model’s output distribution for each input, with the most frequent sample serving as a reference to calculate predictive entropy (PE). Experimental evaluations across six LLMs and four datasets (MedMCQA, MedQA, MMLU, MMLU-Pro) demonstrate that frequency-based PE outperforms logit-based PE in distinguishing between correct and incorrect predictions, as measured by AUROC. Furthermore, the method effectively controls the empirical miscoverage rate under user-specified risk levels, validating that sampling frequency can serve as a viable substitute for logit-based probabilities in black-box scenarios. This work provides a distribution-free model-agnostic framework for reliable uncertainty quantification in MCQA with guaranteed coverage, enhancing the trustworthiness of LLMs in practical applications.
\end{abstract}

\section{Introduction}
LLMs have witnessed substantial progress in their evolutionary trajectory~\cite{chen2025does,Chen_2025_CVPR,rong2025backdoor,zhang2023spot}, transforming human-machine interaction~\cite{bi-etal-2025-llava,bi2025cot,bi2025prism,wang2025ascd}, yet their core flaw "hallucination" is prominent: generating content that appears plausible but lacks a factual basis, such as factual errors like fictional events and scientific conclusions, along with logical contradictions and overconfidence issues~\cite{chakraborty2025hallucination,kuhn2023semantic,qiu2024semantic,wang2025word}. This unreliability severely restricts their application in high-risk fields such as healthcare and finance~\cite{singhal2025toward}.

Conformal prediction (CP)~\cite{angelopoulos2021gentle,angelopoulos2025learn,angelopoulos2024conformalrisk}, a classical frequentist statistical framework grounded in hypothesis testing and exchangeable data theory, aims to deliver reliable confidence intervals and uncertainty quantification for machine learning predictions, relying on the "nonconformity score", quantifying the deviation between predicted samples and calibration data to form prediction sets for new queries through hypothesis testing. CP distinguishes itself from traditional uncertainty quantification methods through distinctive features: providing distribution-free risk control by eschewing distributional assumptions and offering nonasymptotic guarantees; ensuring user-specified coverage probability for prediction sets containing the true value, outperforming point estimates; exhibiting model agnosticism by adapting any pre-trained model from point to interval predictions; and maintaining robustness under model misspecification or distribution shifts~\cite{wang-etal-2025-sconu}. Researchers have explored the application of CP to MCQA tasks with notable results~\cite{ye2024benchmarking,kostumov2024uncertainty}. With the proliferation of LLMs via API services, their "black-box" nature necessitates quantifying output uncertainty~\cite{wang2025sample}. For instance, ConU~\cite{wang2024conu} aligns nonconformity scores strictly with uncertainty conditions for acceptable answers, achieving user-specified coverage guarantees of correctness.

In this study, to quantify the uncertainty of the general prediction of the model, we employed a multiple sampling strategy. Specifically, for each input, we conducted repeated independent samplings of the model's output distribution, with the most frequent sample replacing the traditional maximum-probability (logit-based) point prediction as the reference. This modal-output-referenced uncertainty quantification is intuitively valid: output variability from multiple samplings effectively mirrors the model's predictive confidence for specific inputs. The high concentration of samples around the mode indicates low uncertainty, whereas a higher dispersion corresponds to a higher uncertainty. To verify the effectiveness of the above idea, we designed and conducted a series of experiments focusing on evaluating indicators such as Area Under the Receiver Operating Characteristic Curve (AUROC) and the commonly used uncertainty quantification measure of PE to assess the ability of the proposed method to distinguish between correct and incorrect predictions of the model under different uncertainty thresholds. 

In our experimental investigation, which encompassed six datasets and four models, the results demonstrate that frequency-based Predictive Entropy (PE) generally outperforms its logit-based counterpart in terms of performance. Specifically, in the experimental combination of the Qwen2.5-3B-Instruct model and the MedMCQA dataset, the AUROC value obtained by the frequency-based method is 2\% higher than that achieved by the logit-based method. Meanwhile, under various user-specified risk levels, the frequency-based method can effectively constrain the miscoverage rate. Through systematic experimental verification, we not only confirmed the feasibility of using the modal output as a reference benchmark to quantify the overall uncertainty of the model but also further demonstrated its potential in improving the effectiveness of uncertainty assessment.

\section{Related Work}
\paragraph{Background of CP.}
CP~\cite{angelopoulos2021gentle,wang2025coin}, a salient uncertainty quantification methodology in machine learning, ensures predictive reliability by providing prediction intervals or sets. Unlike conventional methods relying on strict data distribution assumptions, CP operates under the weaker “exchangeability” condition. As a potent and flexible framework, it dispenses with data distribution presuppositions, anchors on data exchangeability, and leverages non - conformity scores and the exchangeability hypothesis to ensure statistical coverage of generated prediction sets—critical for dependable uncertainty quantification—while enabling adaptability to real - world scenarios with tenuous or inapplicable distributional presumptions. 

Recent work focuses on constructing prediction sets by calibrating sampling stopping rules and rejection rules, ensuring the inclusion of at least one acceptable response while maintaining distribution-free performance guarantees. Additionally, another line of research has demonstrated that human-machine collaborative CP can enhance performance in relevant tasks. Furthermore, a separate study has proposed an approach to building distribution-free prediction sets with finite-sample conditional guarantees, effectively bridging the gap between marginal and conditional coverage.

\paragraph{CP in MCQA.}
CP has been extensively employed in MCQA tasks, with its core essence lying in providing uncertainty quantification and reliable performance guarantees for LLMs. Specifically, the prediction sets constructed by CP are capable of including the correct answer with a user-specified coverage probability $\alpha$, and this guarantee is independent of the underlying data distribution or model architecture. Consequently, it is well-suited for LLMs that are costly to retrain or inaccessible through commercial APIs.

In relevant studies, several lines of research have been explored: some works have achieved distribution-free uncertainty quantification, verified the role of CP in selective classification, and evaluated its performance when the exchangeability assumption is violated. Notably, Vishwakarma et al. (2024) proposed an optimized CP method, which enhances the decision quality and final accuracy of MCQ tasks by reprompting LLMs to reduce answer options. Barber et al. (2019) investigated the feasibility of achieving conditional coverage under distribution-free conditions and put forward a variety of relaxation strategies. In addition, Kumar et al. (2023) validated the effectiveness of CP in MCQA tasks through experiments and found that it exhibits favorable calibration across different domains.

\section{Method}
\subsection{Preliminaries}
This paper focuses on the problem of answer prediction for multiple-choice questions. We partition the samples into a calibration set $ \mathcal{D}_{\text{cal}} = \{x_i\}_{i=1}^n $ and a test sample $ x_{n+1} $, where each $ x_i $ corresponds to a true label $ y_i^* $. Given a pre-trained classification model $ F $ with K-classification capability, for any input sample $ x $, $ F(x)_y $ denotes the output score of the model for label $ y $. Let the input question be $ x $, with its valid answer space denoted as $ \mathcal{A} = \{a_1, a_2, \ldots, a_K\} $, where $ y^* \in \mathcal{A} $ represents the true answer. The output of the model for input $ x $ is a probability distribution $ P(y|x) $ over the answer space, characterizing the confidence level of each option. The research aims to enhance the reliability and interpretability of answer prediction by quantifying  predictive entropy and controlling error risks.

\subsection{Conformal Prediction}
\paragraph{Frequency-based PE}
For a specific multiple-choice question, first, $M$ independent samples are performed within the space of its valid answers (ensuring that each sampling result conforms to the question's response specifications, such as constraints on the number of options). From this, a sample set $E$ containing $M$ candidate answers is obtained. Based on this sample set, the frequency $\hat{P}(a)$ of each candidate answer is calculated, and the candidate answer with the highest frequency is selected.
\begin{equation} 
\hat{P}(a)=\frac{1}{M}\mathbb{I}(a^{m}=a)
\end{equation}
Furthermore, based on the empirical frequency distribution constructed from the above $M$ samplings, the frequency-based predictive entropy can be defined as the predictive entropy of this empirical distribution, with its calculation formula as follows:
\begin{equation} 
\text{PE}_{\text{freq}} = -\sum_{i=1}^{n} \hat{P}_i(a) \log_b \hat{P}_i(a)
\end{equation}

\paragraph{Risk Control}
CP is a statistical prediction framework capable of providing finite-sample coverage guarantees. Its core lies in constructing a prediction set that contains the true value, with the probability that this set includes the true value (coverage rate) being no less than the pre-specified confidence level $ 1 - \alpha $. For input from the test set $X$ and the corresponding true label $Y$, if the prediction set is $C(X)$, the coverage rate must satisfy:
\begin{equation} 
\mathbb{P}(Y \in C(X)) \geq 1 - \alpha
\end{equation}
This guarantee must hold under the conditions of finite samples and distribution-freeness. 
For each sample $ x_i $ in the calibration set, we define its non-conformity score as:
\begin{equation} 
s_i = 1 - F(x_i)_{y_i^*} 
\end{equation}
where $ F(x_i)_{y_i^*} $ is the output score of the model for the true label $ y_i^* $.
For the set of non-conformity scores $ \{s_i\}_{i=1}^n $ from the calibration set, after sorting them in ascending order, we calculate the quantile $ \hat{q} $ corresponding to the quantile point $ \frac{|(n+1)(1-\alpha)|}{n} $, that is:
\begin{equation} 
\hat{q} = \text{quantile}\left( \{s_i\}_{i=1}^n, \frac{|(n+1)(1-\alpha)|}{n} \right)
\end{equation}

For the test sample $ (x_{n+1}, y) $, its non-conformity score is defined as:
\begin{equation} 
s_{n+1}(x_{n+1}, y) = 1 - F(x_{n+1})_y
\end{equation}
Based on the above quantile threshold $ \hat{q} $, the prediction set is constructed as follows:
\begin{equation} 
\mathcal{C}(x_{n+1}) = \{ y \mid s_{n+1}(x_{n+1}, y) \leq \hat{q} \}
\end{equation}
Assuming that the $ n $ samples in the calibration set and the test sample $ x_{n+1} $ satisfy the independent and identically distributed (i.i.d.) condition, we have:
\begin{equation} 
\mathbb{P}\left( s_{n+1}(x_{n+1}, y_{n+1}^*) \leq s_i \right) = \frac{i}{n+1}
\end{equation}
Furthermore, the coverage probability of the prediction set satisfies:
\begin{equation} 
\mathbb{P}\left( y_{n+1}^* \in C(x_{n+1}) \right) = \mathbb{P}\left( s_{n+1}(x_{n+1}, y_{n+1}^*) \leq s_{\lfloor (n+1)(1-\alpha) \rfloor} \right) = \frac{\lfloor (n+1)(1-\alpha) \rfloor}{n+1} \geq 1 - \alpha
\end{equation}
That is, the prediction set can guarantee coverage of the true label with a probability of no less than $ 1 - \alpha $.

\begin{table}[!t]
    \caption{Logit vs Frequency Entropy}
    \centering
    \begin{minipage}[b]{0.45\textwidth}
        \centering
        \subcaption[]{MMLU}
        \footnotesize
        \begin{tabular}{lccc}
        \toprule
        Model & Model logit & Sampling set \\
        \midrule
        Qwen2.5-3B-Instruct & 0.6266 & 0.6383 \\
        Llama-3.2-1B & 0.4909 & 0.5082 \\
        Qwen2.5-7B-Instruct & 0.6208 & 0.6311 \\
        Llama-3.1-8B-Instruct & 0.8059 & 0.8215 \\
        Vicuna-7b-v1.5 & 0.6896 & 0.7041 \\
        Vicuna-13b-v1.5 & 0.7298 & 0.7346 \\
        Average & 0.6606 & 0.6729 \\
        \bottomrule
        \end{tabular}
        \end{minipage}
        \hfill
        \begin{minipage}[b]{0.45\textwidth}
        \centering
        \subcaption[]{MEDMCQA}
        \footnotesize
        \begin{tabular}{lccc}
        \toprule
        Model & Model logit & Sampling set \\
        \midrule
        Qwen2.5-3B-Instruct & 0.6262 & 0.6462 \\
        Llama-3.2-1B & 0.5017 & 0.4796 \\
        Qwen2.5-7B-Instruct & 0.6289 & 0.6442 \\
        Llama-3.1-8B - Instruct & 0.7404 & 0.7415 \\
        Vicuna-7b-v1.5 & 0.5928 & 0.6121 \\
        Vicuna-13b-v1.5 & 0.6201 & 0.6314 \\
        Average & 0.6183& 0.6258 \\
        \bottomrule
        \end{tabular}
    \end{minipage}

    \begin{minipage}[b]{0.45\textwidth}
        \centering
        \subcaption[]{MMLU\_PRO}
        \footnotesize
        \begin{tabular}{lccc}
        \toprule
        Model & Model logit & Sampling set \\
        \midrule
        Qwen2.5-3B-Instruct & 0.6457 & 0.6553 \\
        Llama-3.2-1B & 0.5293 & 0.5285 \\
        Qwen2.5-7B-Instruct & 0.6524 & 0.6500 \\
        Llama-3.1-8B-Instruct & 0.7401 & 0.7546 \\
        Vicuna-7b-v1.5 & 0.6630 & 0.6589 \\
        Vicuna-13b-v1.5 & 0.6946 & 0.6998 \\
        Average & 0.6542 & 0.6579 \\
        \bottomrule
        \end{tabular}
        \end{minipage}
        \hfill
        \begin{minipage}[b]{0.45\textwidth}
        \centering
        \subcaption[]{MEDQA}
        \footnotesize
        \begin{tabular}{lccc}
        \toprule
        Model & Model logit & Sampling set \\
        \midrule
        Qwen2.5-3B-Instruct & 0.5669 & 0.5651 \\
        Llama-3.2-1B & 0.4861 & 0.4771 \\
        Qwen2.5-7B-Instruct & 0.5670 & 0.5645 \\
        Llama-3.1-8B-Instruct & 0.7687 & 0.7719 \\
        Vicuna-7b-v1.5 & 0.5725 & 0.5767 \\
        Vicuna-13b-v1.5 & 0.6306 & 0.6299 \\
        Average & 0.5986 & 0.5975 \\
        \bottomrule
        \end{tabular}
    \end{minipage}
\end{table}

\section{Evaluations}
\subsection{Experimental Set-up}
\paragraph{Base LLMs.}
We perform empirical evaluations on six large language models (LLMs) with diverse sizes and architectural designs to ensure a comprehensive analysis. The selected models include Vicuna-7B-v1.5, Vicuna-13B-v1.5, Qwen2.5-3B-Instruct, Qwen2.5-7B-Instruct, Llama-3.1-8B-Instruct, and Llama-3.2-1B. For all open-source LLMs, we adopt the default generation configurations and checkpoints available through the HuggingFace platform.
\paragraph{Datasets.}
To evaluate the efficacy of the proposed method and validate its correctness coverage guarantees, we employ four datasets: MedMCQA (a large-scale multi-subject, multi-choice medical question-answering dataset), MedQA (a large-scale open-domain dataset sourced from medical examinations), MMLU (a comprehensive cross-disciplinary benchmark for assessing LLMs’ general knowledge and problem-solving abilities), and MMLU-Pro (an enhanced MMLU variant with more challenging questions to probe the depth of models’ understanding and reasoning).

\begin{figure}[!t]
    \centering
    \begin{minipage}[b]{\textwidth}
        \begin{minipage}[b]{0.24\textwidth}
            \includegraphics[width=\linewidth]{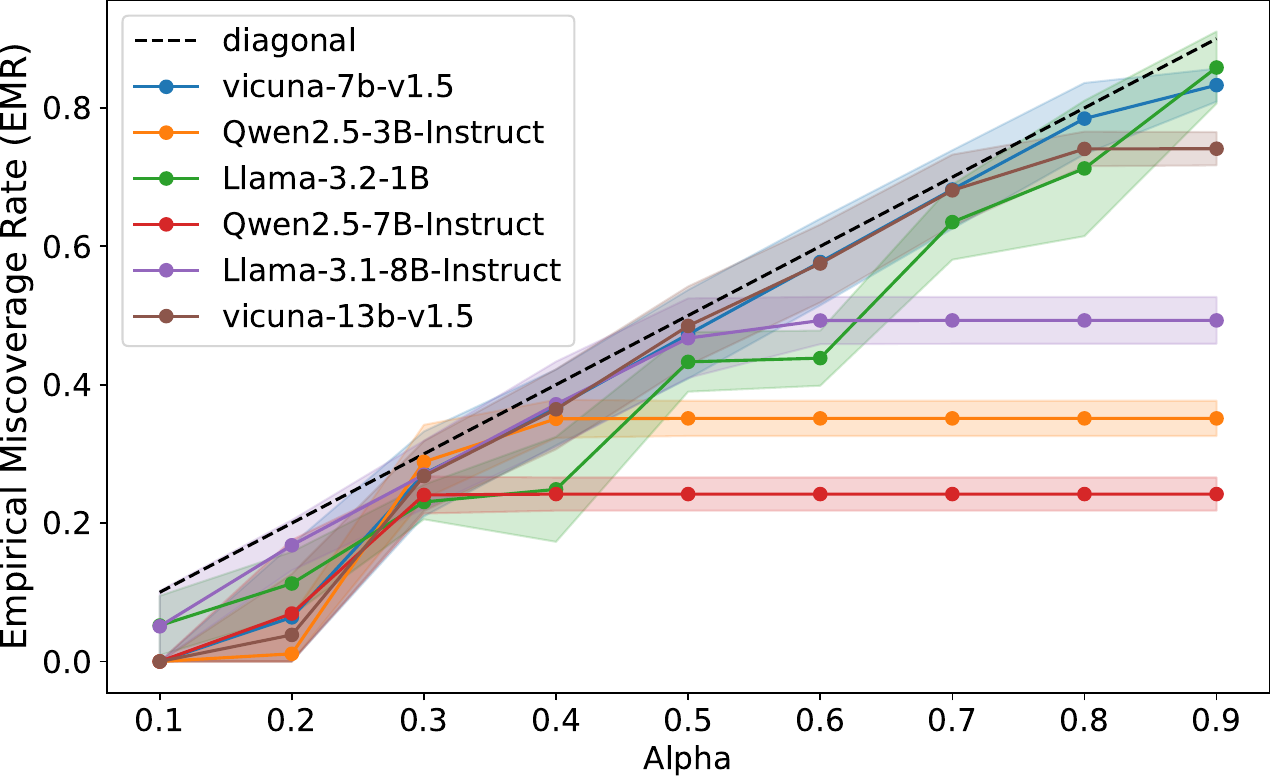}
            \subcaption{Clinical Knowledge}
        \end{minipage}
        \hspace{0pt}
        \begin{minipage}[b]{0.24\textwidth}
            \includegraphics[width=\linewidth]{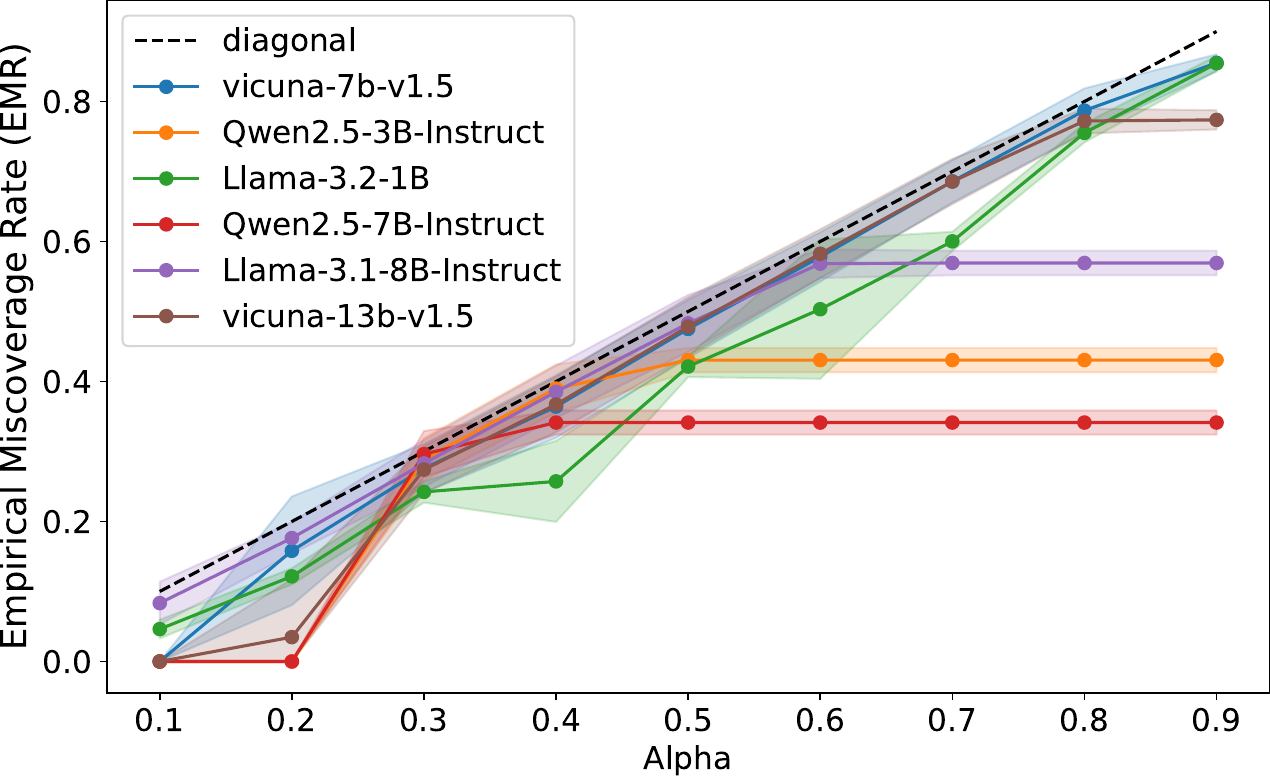}
            \subcaption{College Chemistry}
        \end{minipage}
        \hspace{0pt}
        \begin{minipage}[b]{0.24\textwidth}
            \includegraphics[width=\linewidth]{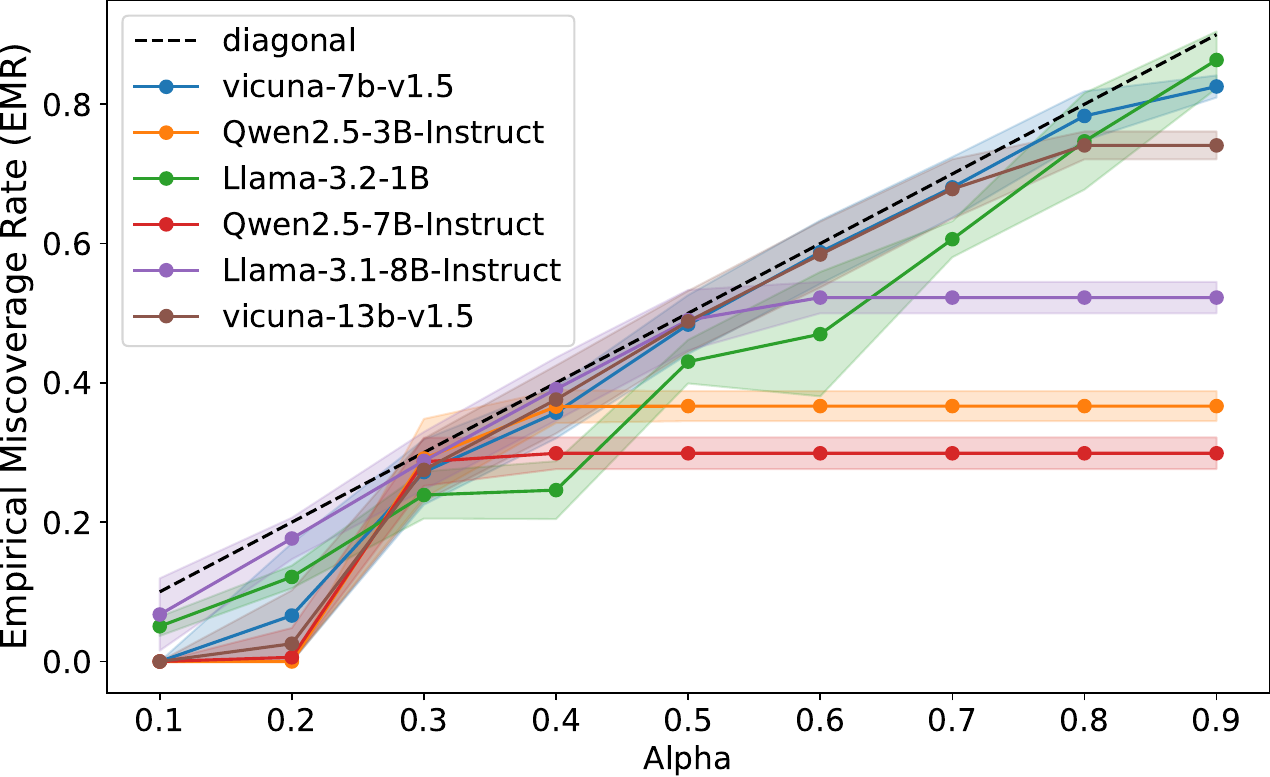}
            \subcaption{College Medicine}
        \end{minipage}
        \hspace{0pt}
        \begin{minipage}[b]{0.24\textwidth}
            \includegraphics[width=\linewidth]{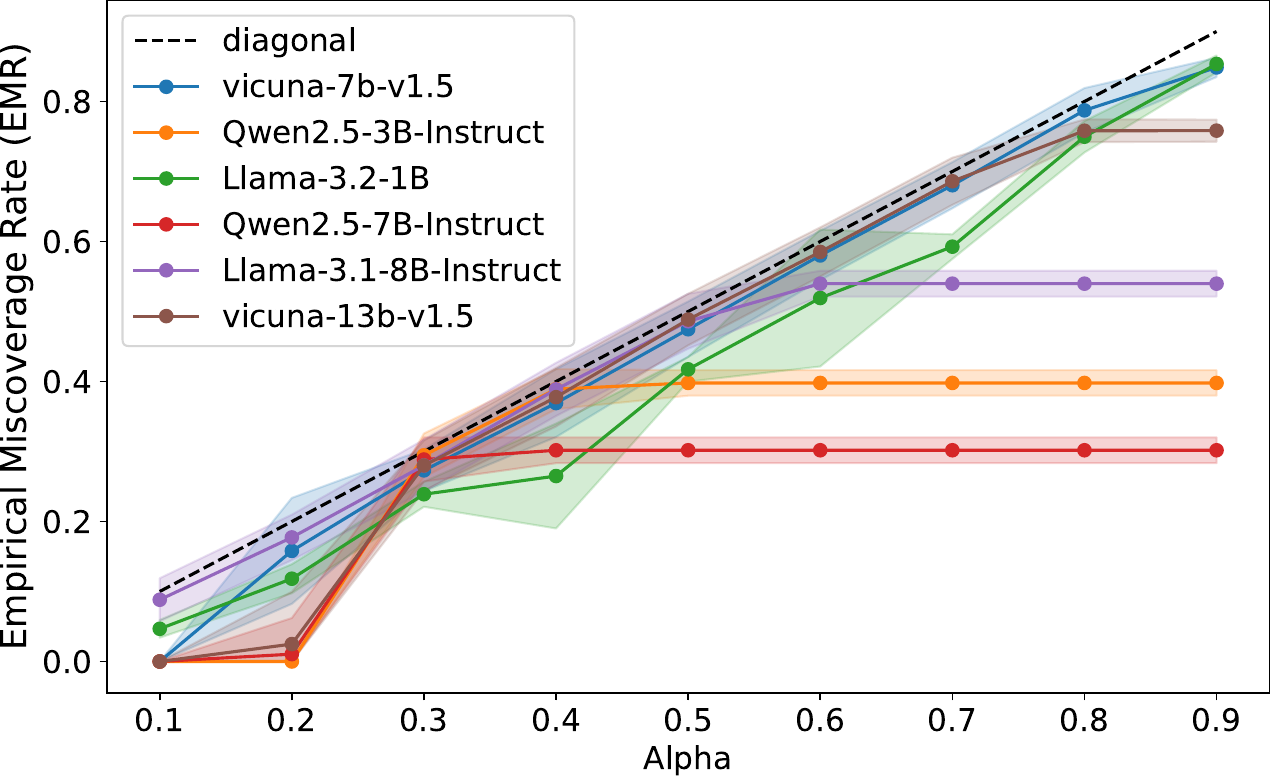}
            \subcaption{Professional Medicine}
        \end{minipage}
    \end{minipage}
    \begin{minipage}[b]{\textwidth}
        \begin{minipage}[b]{0.24\textwidth}
            \includegraphics[width=\linewidth]{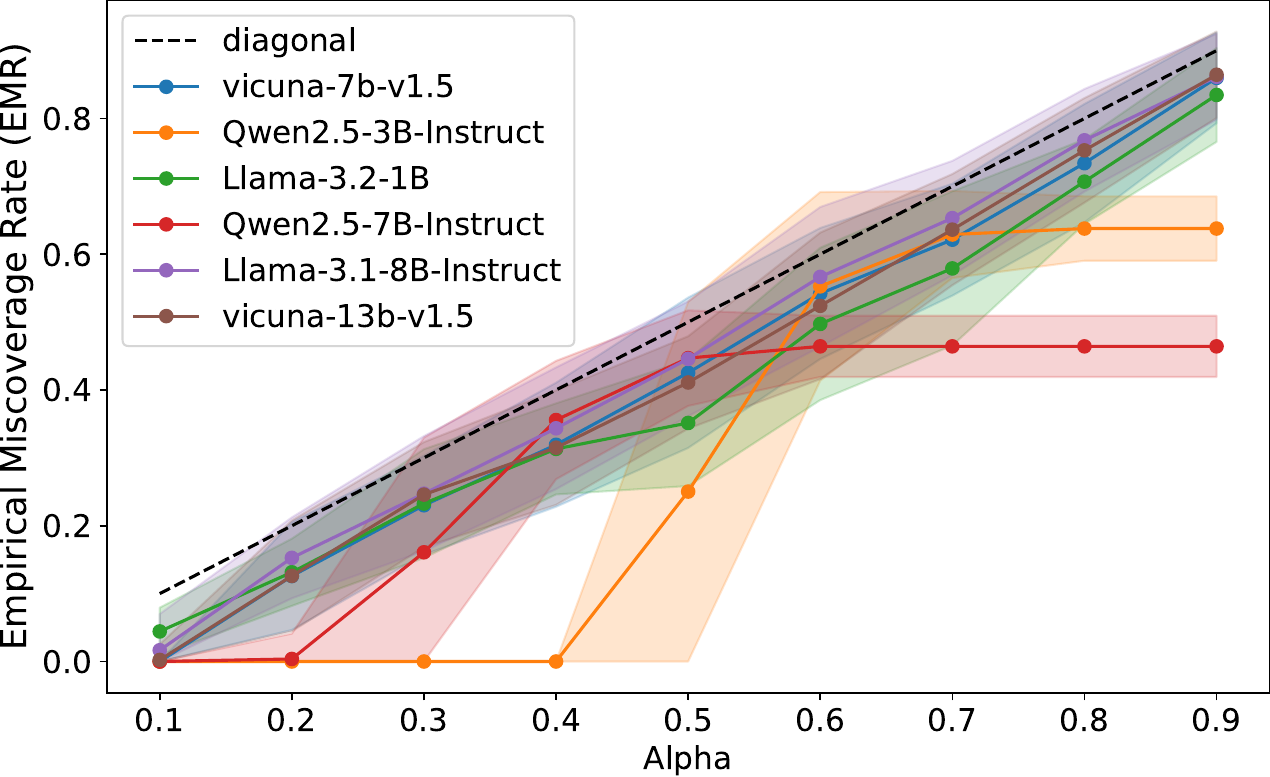}
            \subcaption{Machine Learning}
        \end{minipage}
        \hspace{0pt}
        \begin{minipage}[b]{0.24\textwidth}
            \includegraphics[width=\linewidth]{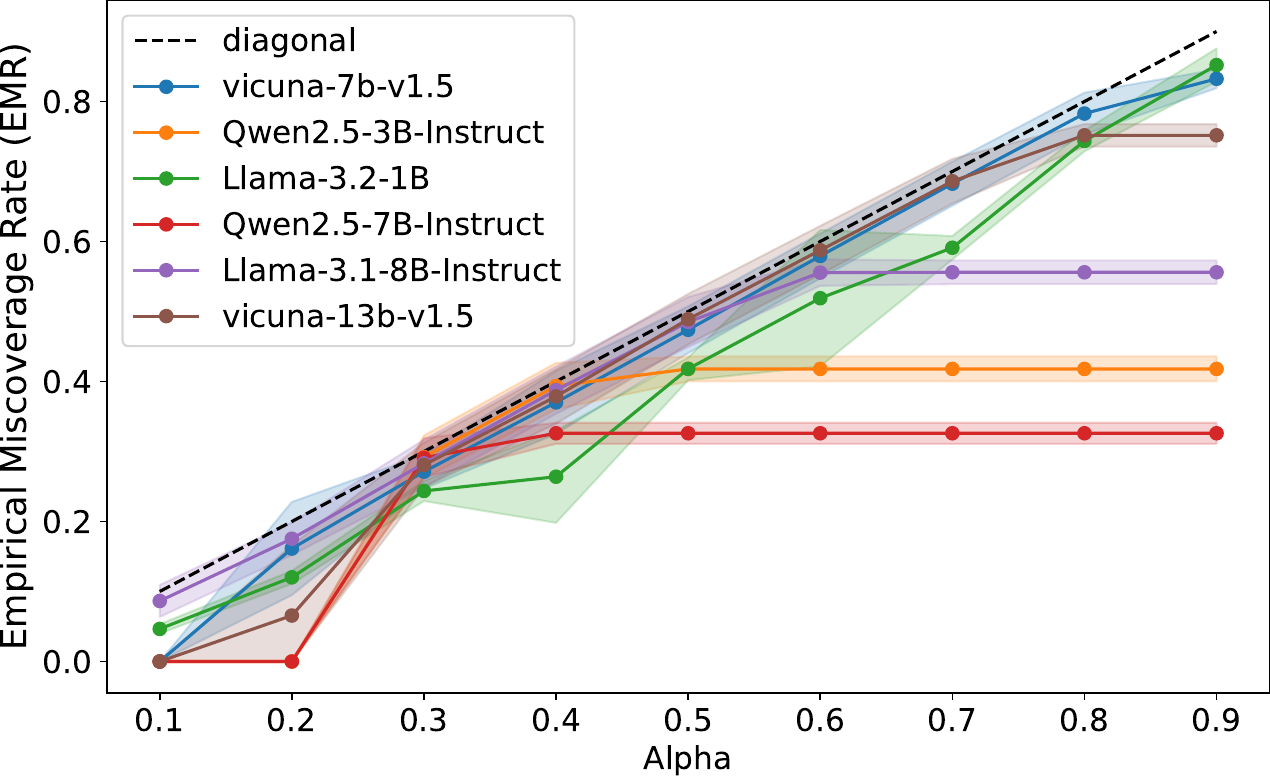}
            \subcaption{Computer Security}
        \end{minipage}
        \hspace{0pt}
        \begin{minipage}[b]{0.24\textwidth}
            \includegraphics[width=\linewidth]{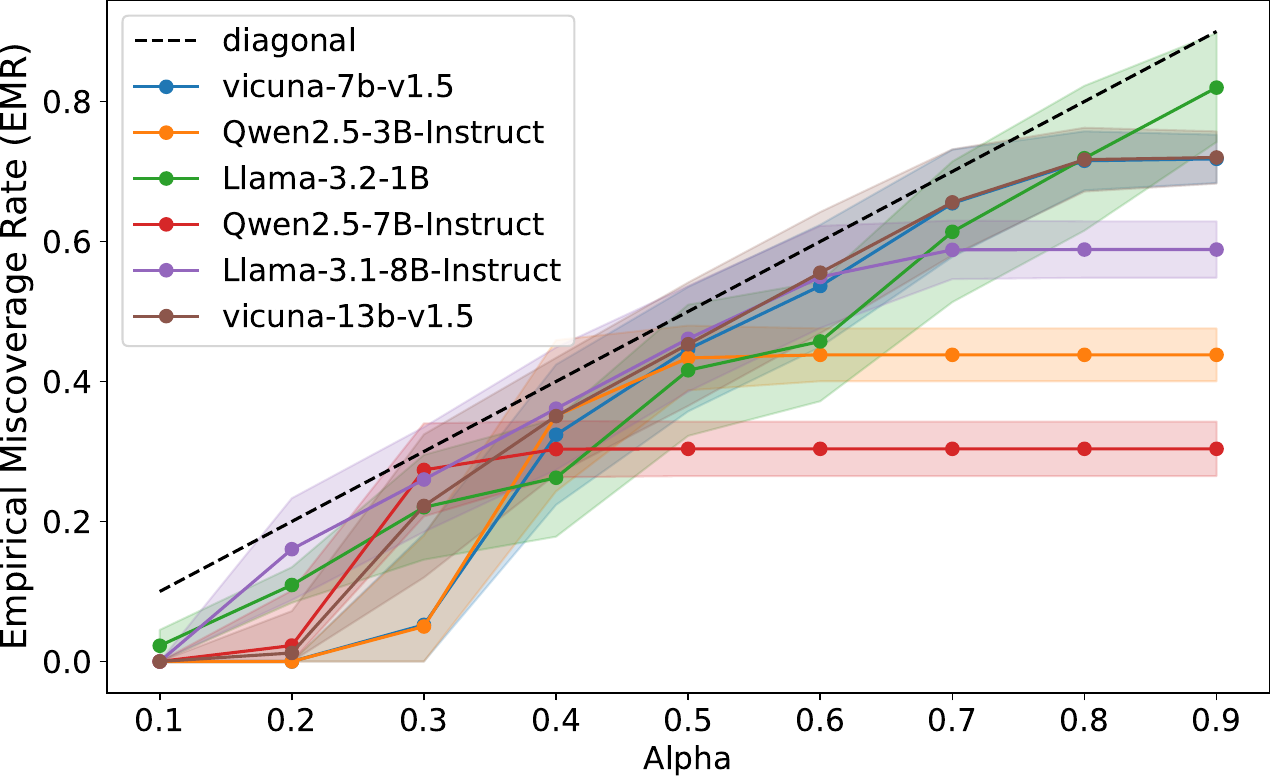}
            \subcaption{Anatomy}
        \end{minipage}
        \hspace{0pt}
        \begin{minipage}[b]{0.24\textwidth}
            \includegraphics[width=\linewidth]{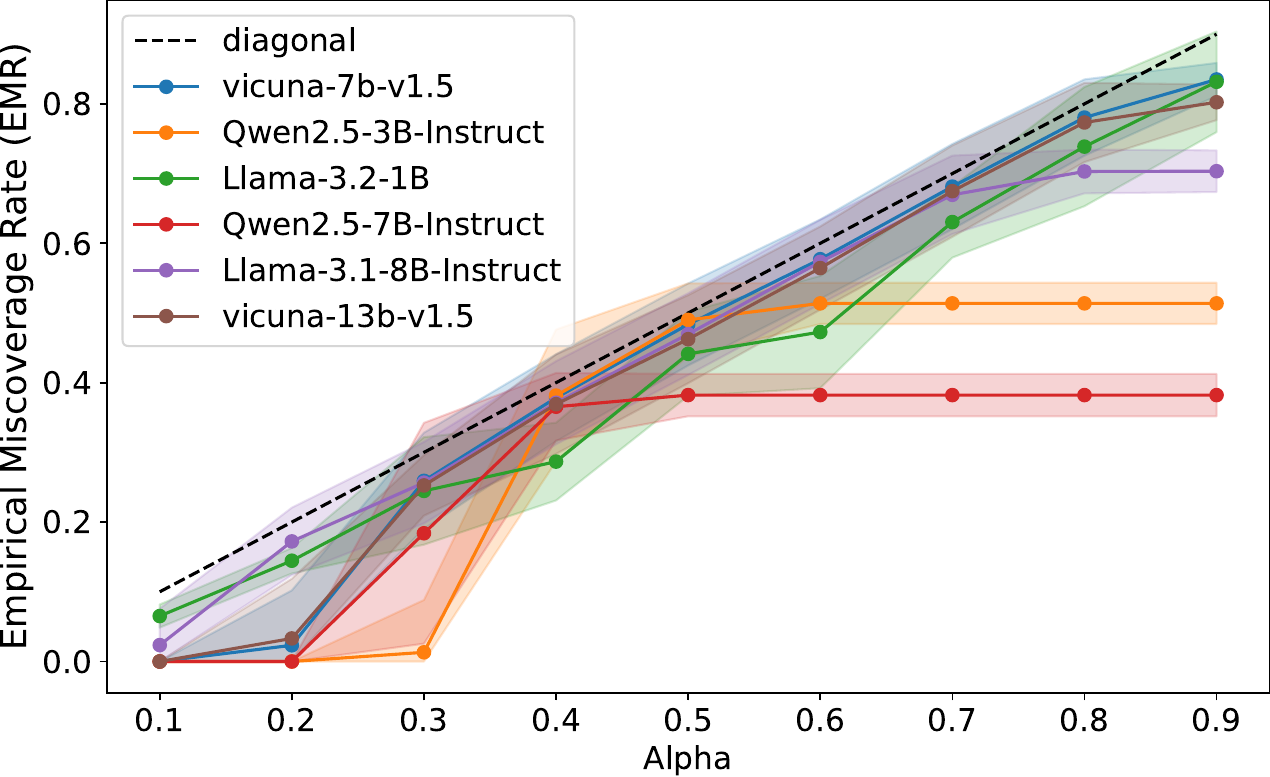}
            \subcaption{Formal Logic}
        \end{minipage}
    \end{minipage}
    \caption{Empirical Miscoverage Rate on MMLU}
\end{figure}

\begin{figure}[!t]
    \centering
    \begin{minipage}[b]{\textwidth}
        \begin{minipage}[b]{0.24\textwidth}
            \includegraphics[width=\linewidth]{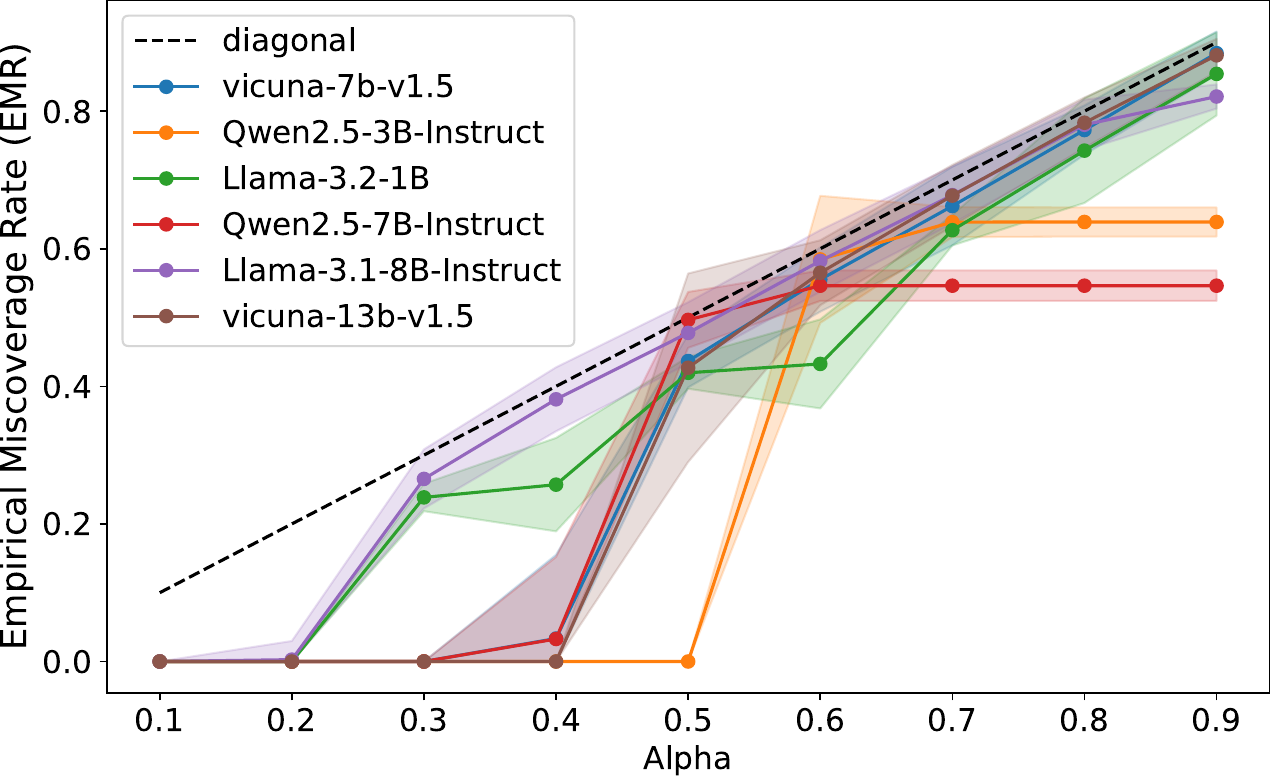}
            \subcaption{Health}
        \end{minipage}
        \hspace{0pt}
        \begin{minipage}[b]{0.24\textwidth}
            \includegraphics[width=\linewidth]{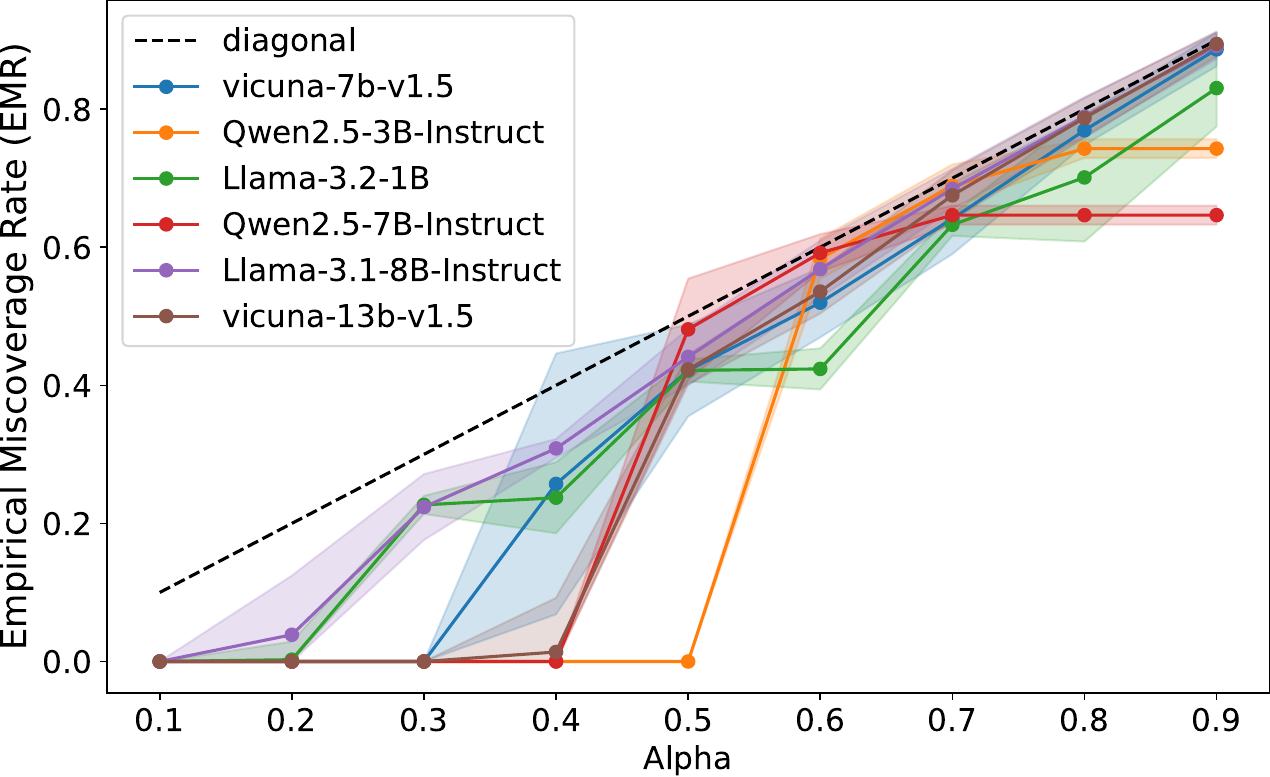}
            \subcaption{Math}
        \end{minipage}
        \hspace{0pt}
        \begin{minipage}[b]{0.24\textwidth}
            \includegraphics[width=\linewidth]{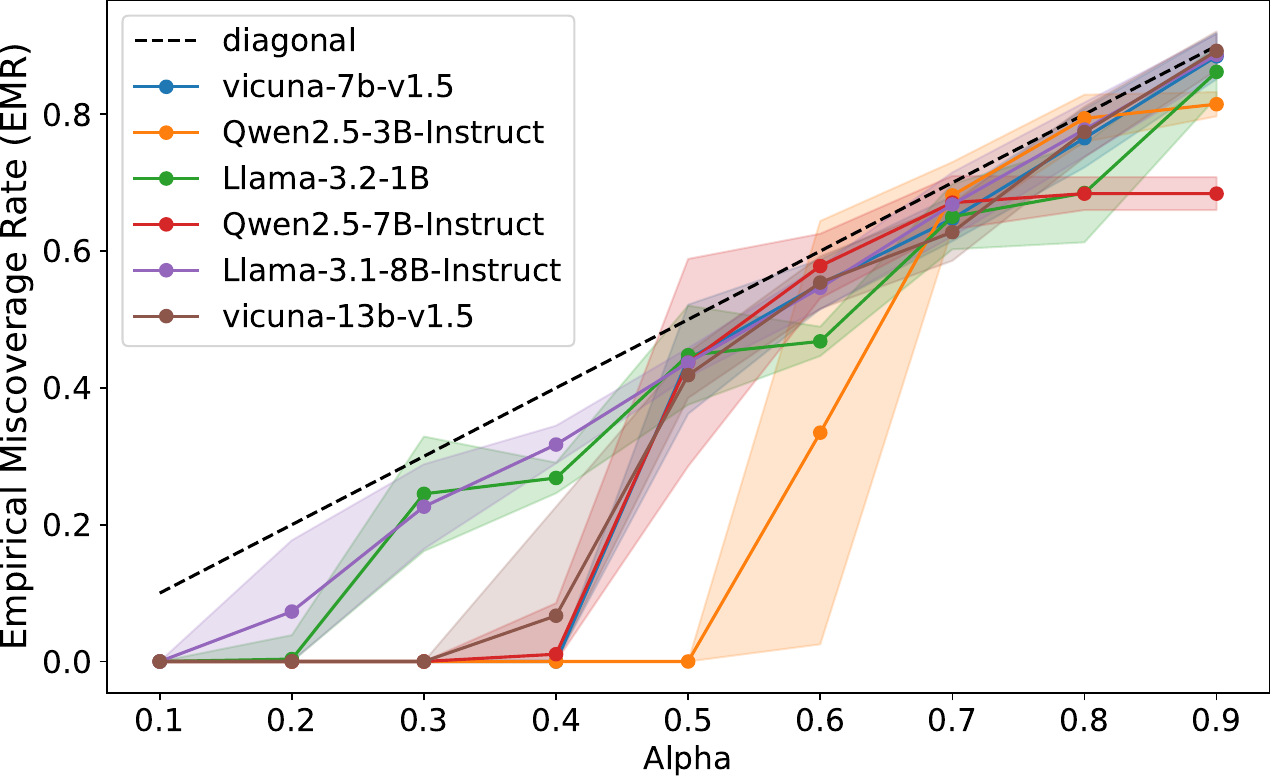}
            \subcaption{Business}
        \end{minipage}
        \hspace{0pt}
        \begin{minipage}[b]{0.24\textwidth}
            \includegraphics[width=\linewidth]{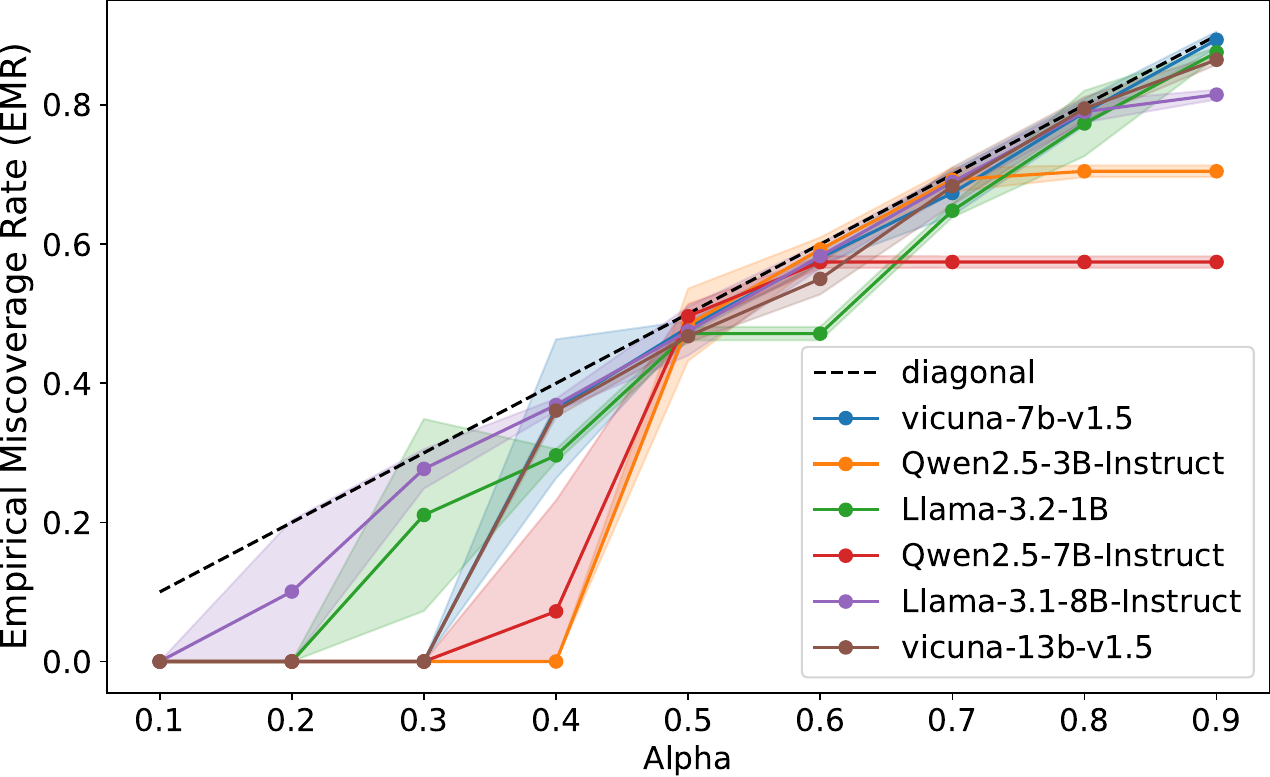}
            \subcaption{History}
        \end{minipage}
    \end{minipage}
    \begin{minipage}[b]{\textwidth}
        \begin{minipage}[b]{0.24\textwidth}
            \includegraphics[width=\linewidth]{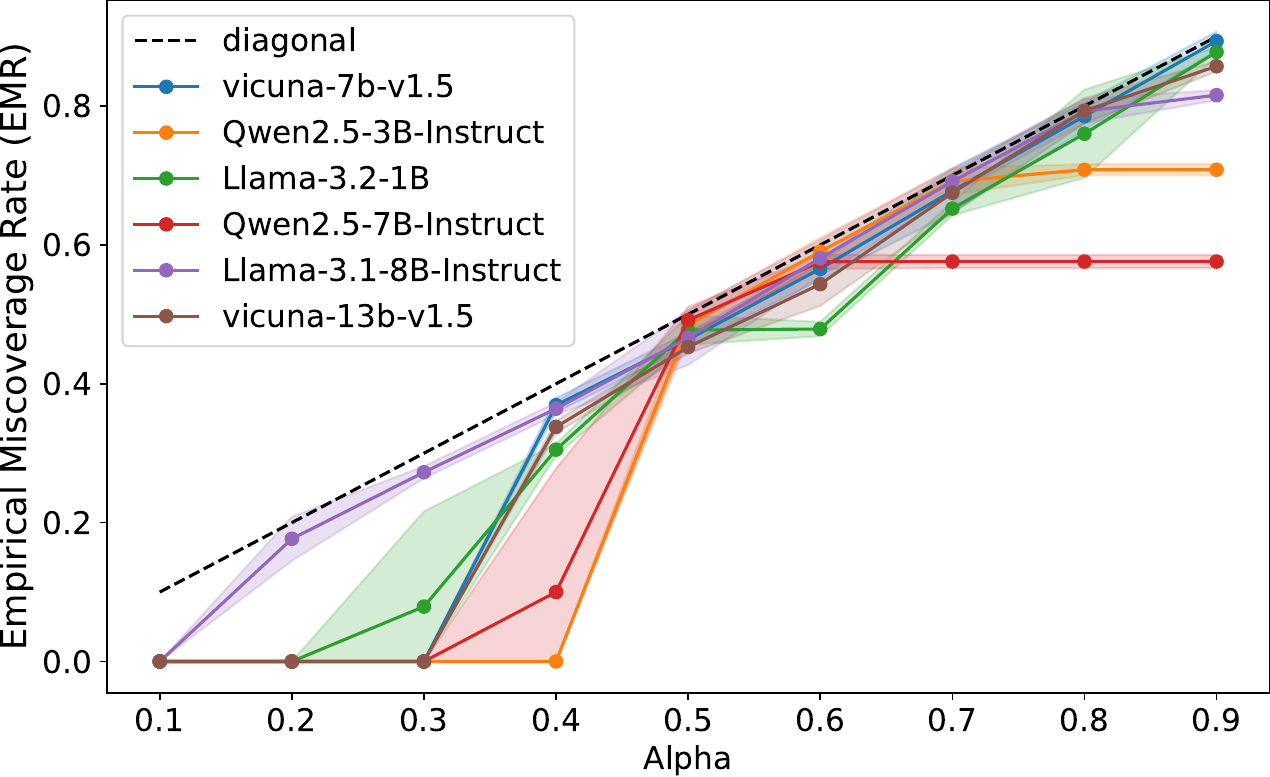}
            \subcaption{Chemistry}
        \end{minipage}
        \hspace{0pt}
        \begin{minipage}[b]{0.24\textwidth}
            \includegraphics[width=\linewidth]{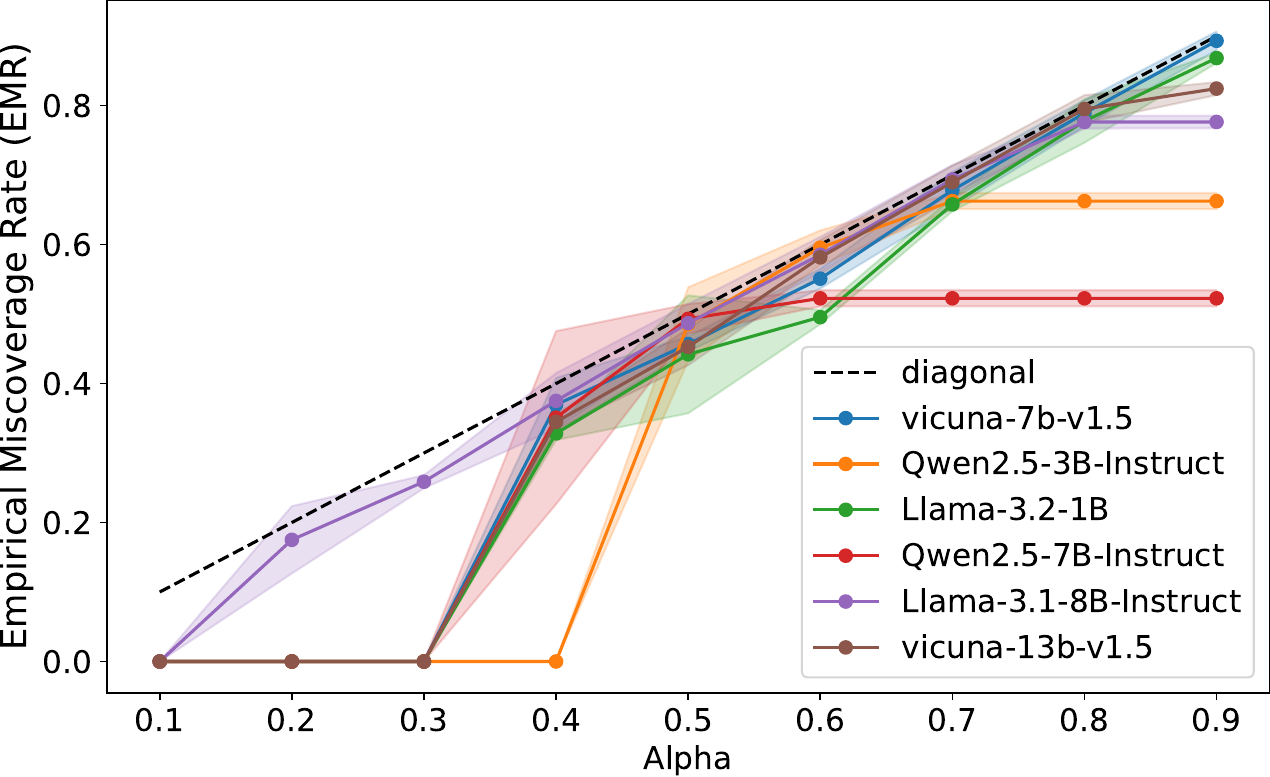}
            \subcaption{Biology}
        \end{minipage}
        \hspace{0pt}
        \begin{minipage}[b]{0.24\textwidth}
            \includegraphics[width=\linewidth]{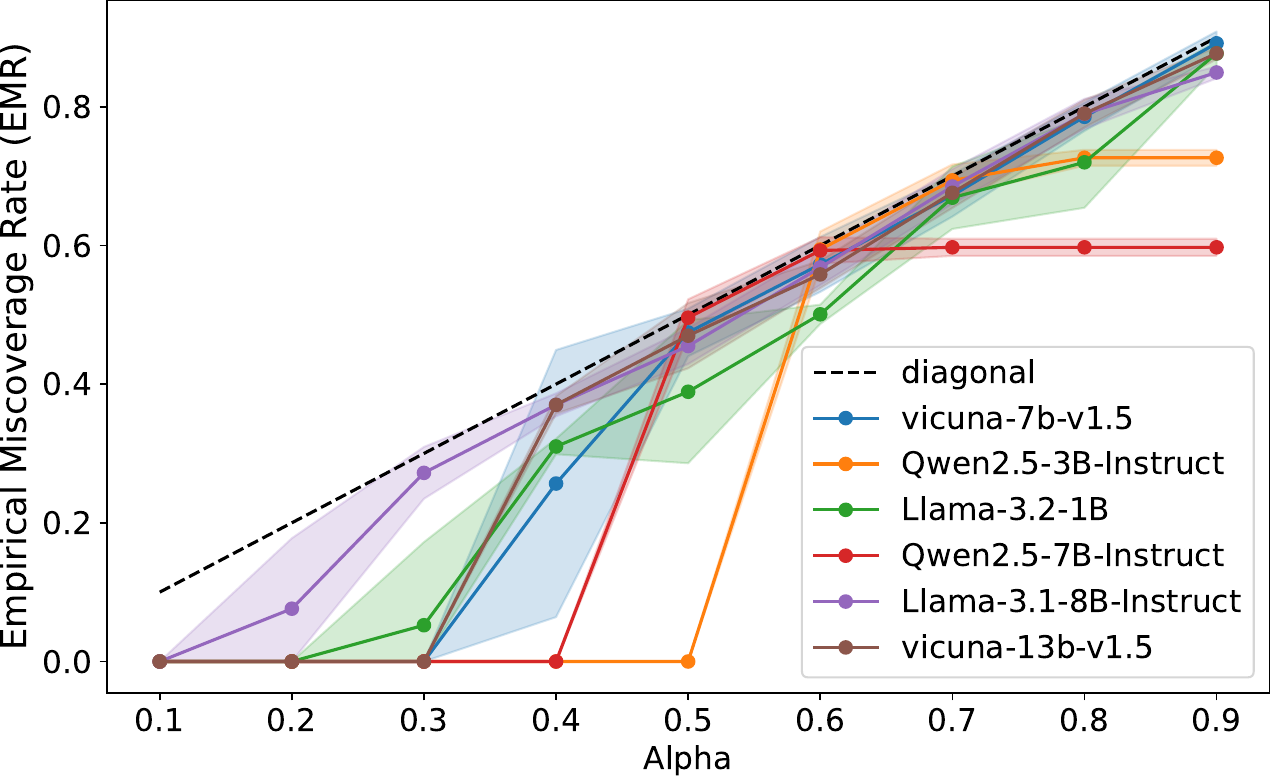}
            \subcaption{Psychology}
        \end{minipage}
        \hspace{0pt}
        \begin{minipage}[b]{0.24\textwidth}
            \includegraphics[width=\linewidth]{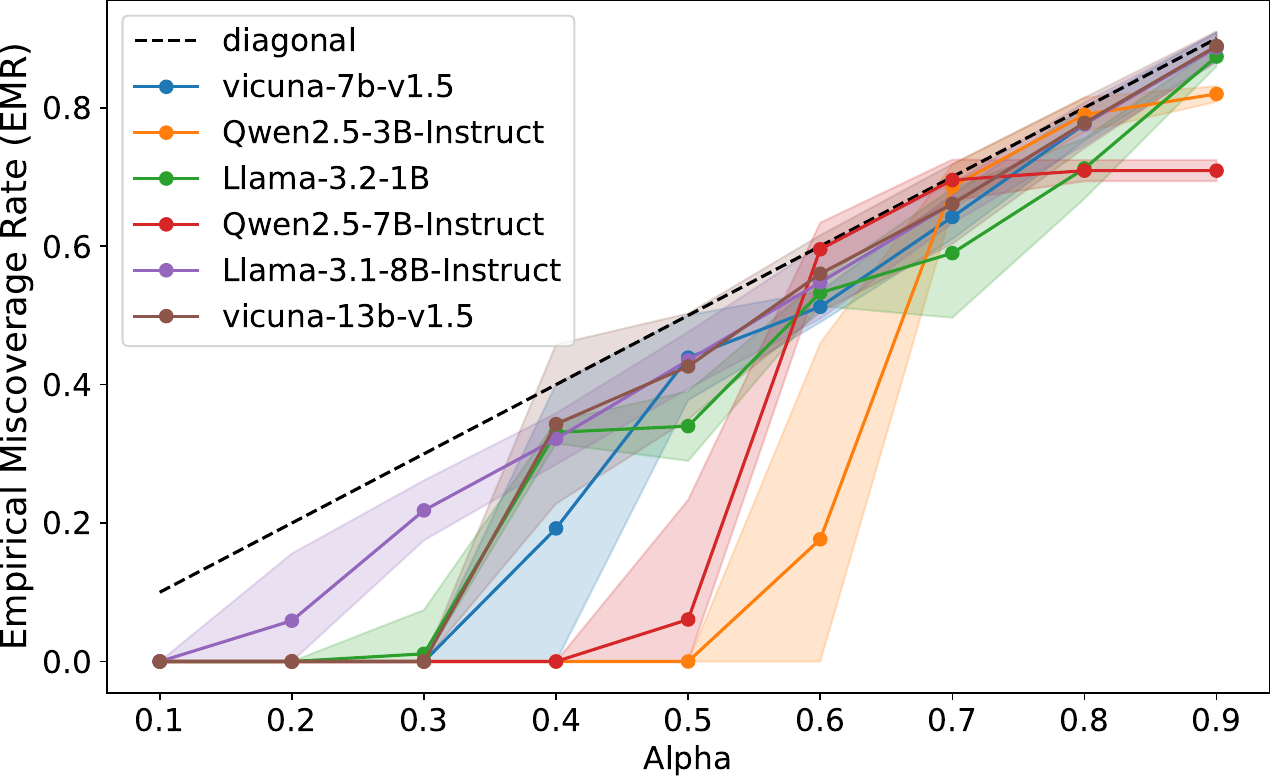}
            \subcaption{Law}
        \end{minipage}
    \end{minipage}
    \caption{Empirical Miscoverage Rate on MMLU Pro}
\end{figure}

\paragraph{Evaluation Metric.}
Consistent with prior literature, we assess uncertainty quantification (UQ) performance by framing it as a task of predicting the trustworthiness of model generations, employing the AUROC to quantify the discriminative power of uncertainty scores in distinguishing between correct and incorrect outputs. For validity verification, we compute the empirical miscoverage rate (EMR) -defined as the proportion of test samples where the prediction set fails to include the correct answer - to ensure it adheres to user-specified thresholds. Additionally, we report the average prediction set size (APSS) as a metric of efficiency, where smaller sizes denote enhanced practical utility.
\paragraph{Hyperparameters.}
We set the maximum length of generation based on dataset characteristics: 36 tokens for CoQA and TriviaQA, and 1 token for MMLU-Pro, MMLU, MedMCQA, and MedQA, to align with the nature of their outputs . We randomly sample 20 candidate generations per prompt for uncertainty quantification, using a temperature of 1.0 for sampling to ensure diverse outputs . The ratio of calibration to test set is set to 0.5 by default, balancing the need for sufficient calibration data and reliable test evaluation. 

\subsection{Sampling Frequency as Logit Substitute for Black-Box LLMs UQ}
In black-box LLMs scenarios where internal logits are inaccessible, UQ must rely on observable outputs. This study investigates whether sampling frequency can substitute logit-based probabilities (via logits + softmax) as a proxy for predictive entropy, enabling logit-free UQ. Correctness is determined by the consistency between the most frequent output and the reference answer; effective uncertainty metrics should distinguish correct from incorrect outputs, aligning with failure prediction criteria.

Predictive entropy for uncertainty scoring is computed from two probability sources: (1) logit-based probabilities in white-box settings (logits + softmax); (2) frequency-based probabilities in black-box settings (derived from the most frequent answer across multiple samplings, inspired by self-consistency theory). Both metrics are evaluated using AUROC to assess their ability to differentiate correct and incorrect outputs, verifying whether frequency-based predictive entropy—using only sampled outputs—achieves AUROC comparable to logit-based methods, thus validating sampling frequency as a viable logit substitute for UQ in black-box LLMs.

Experiments show that frequency-based methods yield AUROC values essentially equivalent to logit-based counterparts in Table 1. For example, across six models on the MMLU dataset, the frequency-based approach achieves a 1.2\% higher average AUROC, with a 0.7\% average increase on MEDMCQA. These results confirm the feasibility of sampling frequency as a logit substitute for UQ in black-box LLMs.

\subsection{Conformal Prediction Coverage}
In this section, based on the theoretical analysis in Equation (9), we analyze the EMR and APSS to validate the performance of the constructed calibrated prediction sets.
\paragraph{Emperical miscoverage rate}
In this section, based on the theoretical analysis presented in Equation (9), we conduct a systematic verification of the Empirical Miscoverage Rate (EMR). Specifically, relying on the conformal framework, we partition the data into a calibration set and a test set at a 1:1 ratio. To ensure the robustness and reliability of the experimental results, for eight subject categories including Clinical Knowledge, College Chemistry, and College Medicine, we independently repeated the partitioning process of the calibration and test sets 100 times for each category. For each subject category, we calculated the mean and standard deviation (mean±std) of the EMR. The experimental results show that the mean EMR of all subject categories is strictly controlled within the user-specified alpha threshold. This outcome verifies the effective control of the conformal framework over the miscoverage rate in practical applications, which is consistent with theoretical expectations.

\begin{figure}[!t]
    \centering
    \begin{minipage}[b]{0.49\textwidth}
        \includegraphics[width=\linewidth]{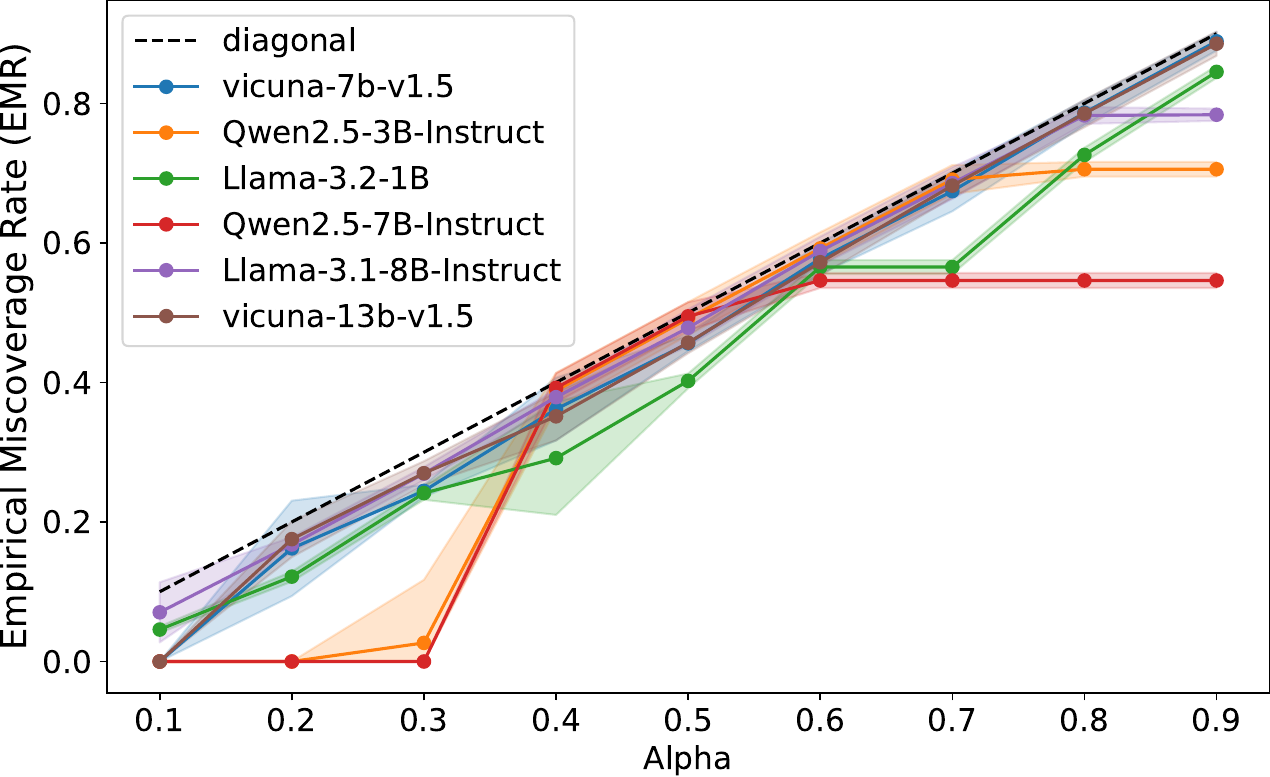}
        \subcaption{MEDMCQA}
    \end{minipage}
    \begin{minipage}[b]{0.49\textwidth}
        \includegraphics[width=\linewidth]{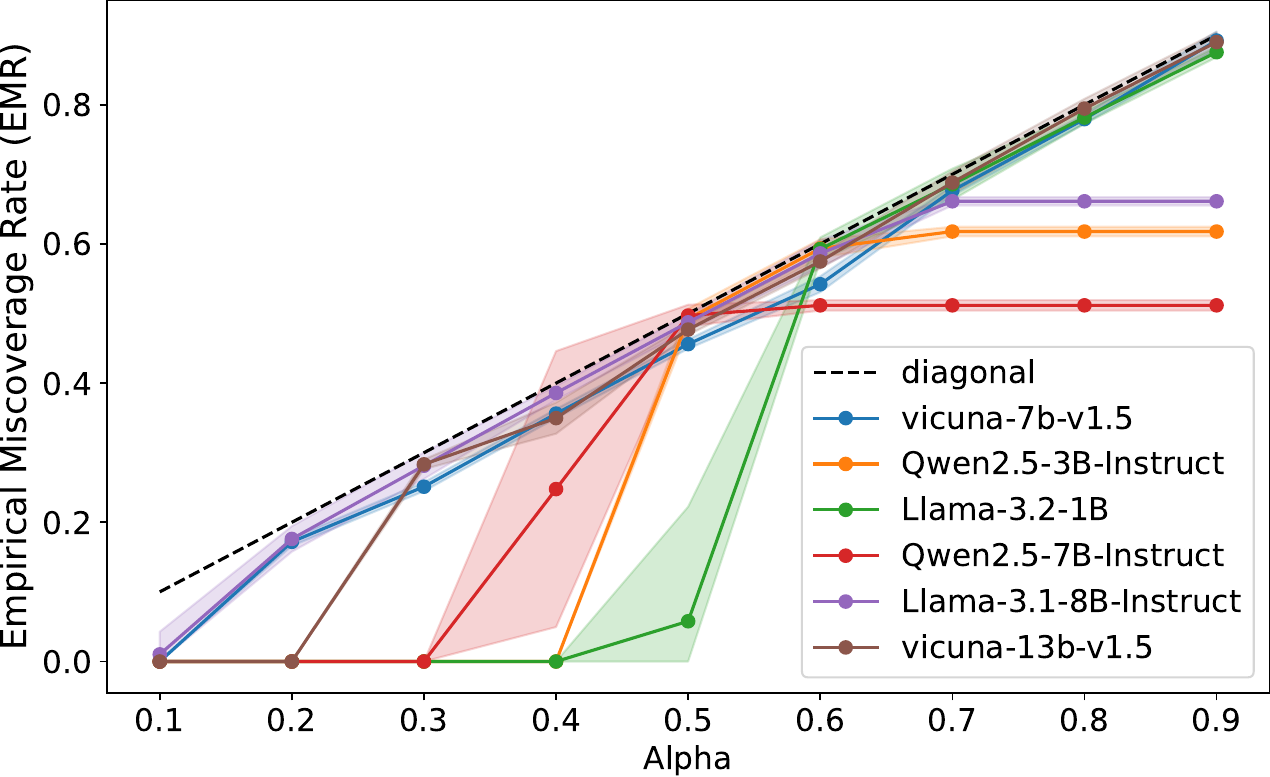}
        \subcaption{MEDQA}
    \end{minipage}
    \caption{Empirical Miscoverage Rate}
\end{figure}

\begin{figure}[!t]
    \centering
    \begin{minipage}[b]{0.49\textwidth}
        \includegraphics[width=\linewidth]{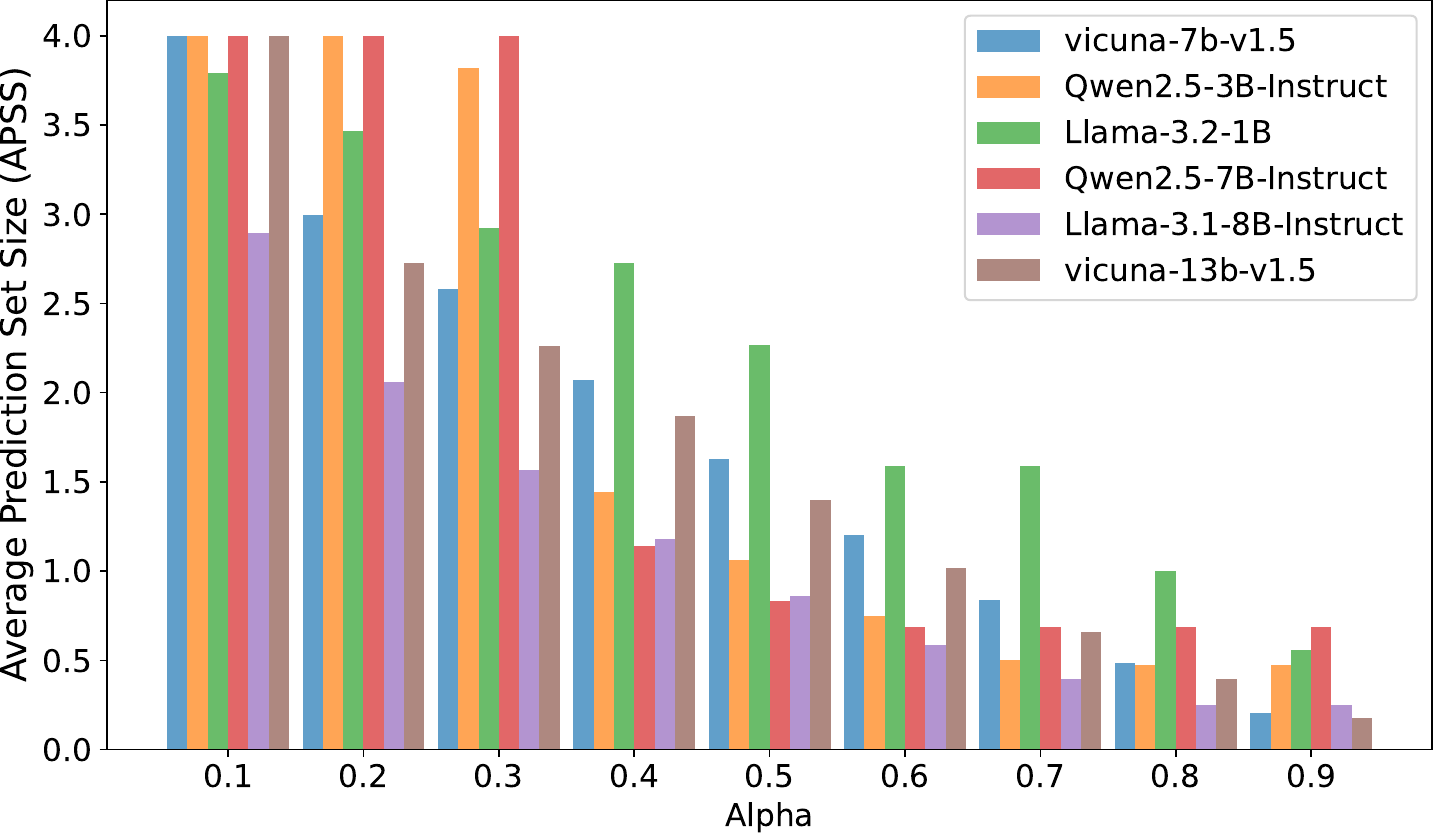}
        \subcaption{MEDMCQA}
    \end{minipage}
    \begin{minipage}[b]{0.49\textwidth}
        \includegraphics[width=\linewidth]{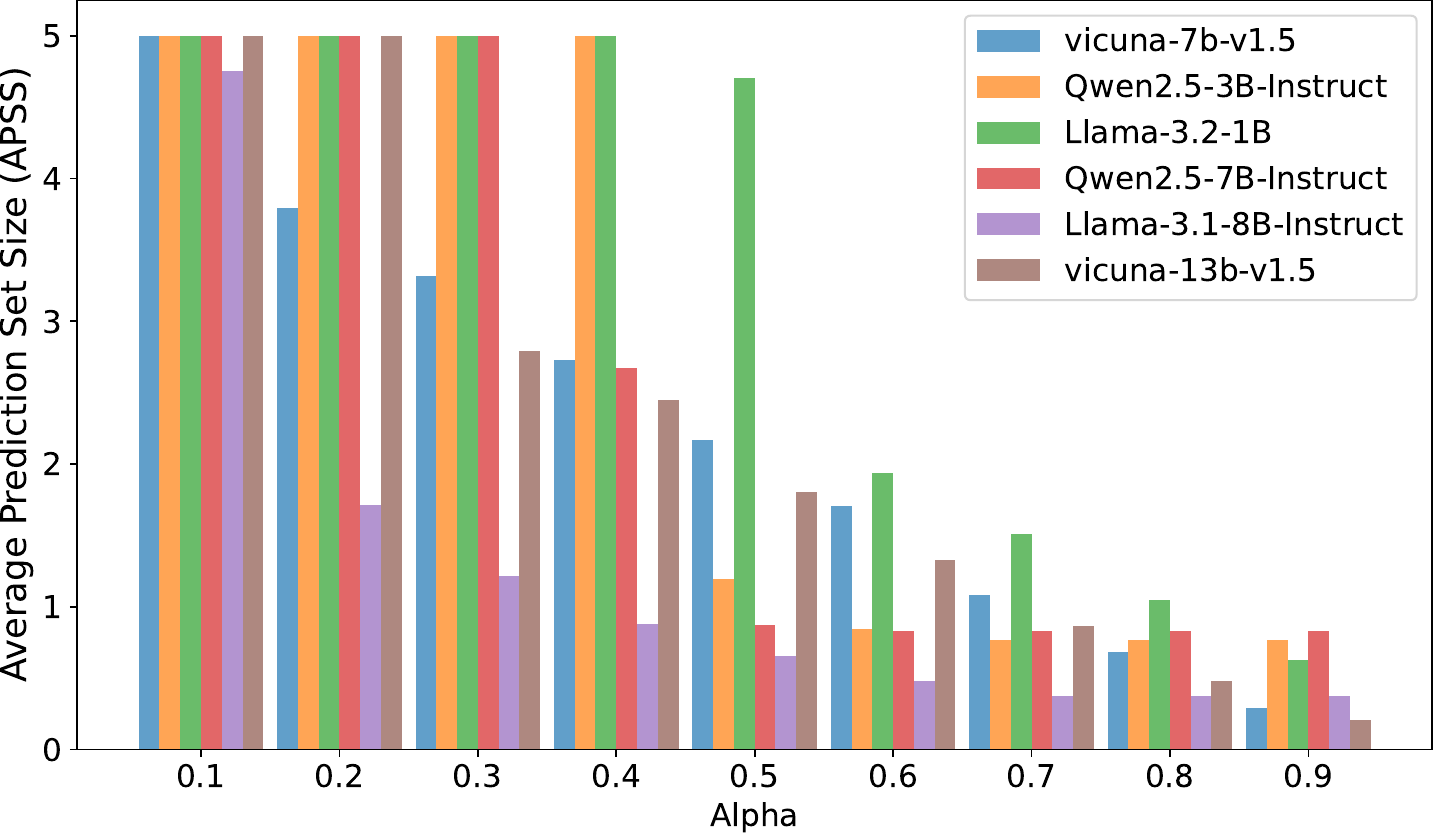}
        \subcaption{MEDQA}
    \end{minipage}
    \caption{Average Prediction Set Size}
\end{figure}

\paragraph{Average prediction set size}
Building on this, we further investigated the dynamic variation pattern of Prediction Set Size with respect to the user-specified alpha coverage threshold. Experimental results demonstrate that as the requirement for coverage guarantees gradually increases, the size of the prediction set exhibits a significant upward trend. This phenomenon is highly consistent with intuitive cognition: the expansion of the prediction set implies a corresponding increase in the probability of containing the correct answer, thereby meeting higher coverage requirements.
Notably, under the same level of coverage guarantee, there exists a clear negative correlation between model performance and prediction set size — superior-performing models can achieve the equivalent coverage target with a smaller prediction set, which directly reflects lower uncertainty in their output results and higher decision-making reliability. This finding not only verifies the intrinsic connection between model performance and the compactness of prediction sets but also provides empirical support for measuring model uncertainty through set size.

\section{Conclusion}

This study focuses on the problem of uncertainty quantification for LLMs in MCQA tasks under black-box settings. It proposes a frequency-based Predictive Entropy (PE) method, which quantifies uncertainty by taking the most frequent sample as a reference through multiple independent samplings of the model's output distribution. Combined with the CP framework, this method constructs prediction sets aiming to ensure that the prediction sets have provable coverage guarantees.

Experimental results demonstrate that across six models and four datasets (MedMCQA, MedQA, MMLU, and MMLU-Pro), the frequency-based PE generally outperforms the logit-based method in terms of uncertainty quantification performance as measured by AUROC. Moreover, it can effectively control the empirical miscoverage rate under various user-specified risk levels. This confirms that sampling frequency can serve as a viable alternative to logit-based probabilities in black-box LLMs, providing an effective means of uncertainty quantification for scenarios where internal logits are inaccessible.

\bibliographystyle{plain}
\bibliography{ref}

\end{document}